\documentclass[10 pt, conference]{ieeeconf}
\IEEEoverridecommandlockouts                              

\usepackage{amsthm}
\usepackage{cite}
\usepackage{amsmath,amssymb,amsfonts}

\usepackage{graphicx}

\usepackage{textcomp}
\usepackage{xcolor}

\usepackage{subfigure}
\usepackage{hyperref}
\usepackage[ruled]{algorithm2e} 
\usepackage{algorithmic}
\usepackage{booktabs}
\usepackage{balance}
\def\BibTeX{{\rm B\kern-.05em{\sc i\kern-.025em b}\kern-.08em
    T\kern-.1667em\lower.7ex\hbox{E}\kern-.125emX}}
    \newtheorem{theorem}{Theorem}

\newtheorem{problem}{Problem}

\newtheorem{definition}{Definition}

\usepackage{threeparttable}

\title{\LARGE \bf
Configuration-Aware Safe Control for Mobile Robotic Arm \\with Control Barrier Functions
\vspace{-10pt}
}

\author{Fan Ding$^{1}$, Han Wang$^{2}$, Jianping He$^{1}$, Yi Ren$^{3}$, and Yu Zheng$^{3}$ 
\thanks{$^{1}$The
Department of Automation, Shanghai Jiao Tong University, Shanghai, China. E-mails: {\tt\small
\{ding0106,jphe\}@sjtu.edu.cn
}}
\thanks{$^{2}$The Department of Engineering Science, University of Oxford, Oxford, United Kingdom. E-mail: {\tt\small
\{han.wang\}@eng.ox.ac.uk
}}
\thanks{$^{3}$The Tencent Robotics X Lab, Shenzhen, China. E-mails: {\tt\small
\{evanyren, petezheng\}@tencent.com
}}
}

\begin{document}

\maketitle
\thispagestyle{empty}
\pagestyle{empty}
\begin{abstract}
Collision avoidance is a widely investigated topic in robotic applications. When applying collision avoidance techniques to a mobile robot, how to deal with the spatial structure of the robot still remains a challenge. 
In this paper, we design a configuration-aware safe control law by solving a Quadratic Programming (QP) with designed Control Barrier Functions (CBFs) constraints, which can safely navigate a mobile robotic arm to a desired region while avoiding collision with environmental obstacles.
The advantage of our approach is that it correctly and in an elegant way incorporates the spatial structure of the mobile robotic arm. This is achieved by merging geometric restrictions among mobile robotic arm links into CBFs constraints.
Simulations on a rigid rod and the modeled mobile robotic arm are performed to verify the feasibility and time-efficiency of proposed method. Numerical results about the time consuming for different degrees of freedom illustrate that our method scales well with dimension.


\end{abstract}

\section{Introduction}
Mobile robotic arm, consisting of robotic arms and mobile or drone bases
have gained popularity in robotics community for their wide applications in navigating and grasping tasks with obstacle avoidance requirements. 
Mobile robotic arms allow more flexible operations to complete assigned tasks and, on the other hand, introduce higher computational complexity because of their spatial structures.
Various techniques have been proposed in the past decades for mobile robots collision avoidance problems, including Cell Decomposition \cite{lingelbach2004path}, Potential Field Methods \cite{tang2010novel}) and heuristic methods (Fuzzy Logic Controller Technique \cite{hong2012application}, Neural Network Technique \cite{tai2016deep}). The readers are referred to \cite{pandey2017mobile} for an overview in this area. When applying these collision avoidance techniques to a mobile robotic arm, how to deal with its spatial structure still remains a challenge. 

Existing collision avoidance algorithms mainly use geometric envelopes to simplify the spatial structure of the robot and reduce the complexity of the optimization.
\cite{zeng2013mobile} enveloped the mobile robot and moving obstacles by spheres with different radii in and a humanoid torso robot which is abstracted as a cylinder to navigate among modeled pedestrians was proposed in \cite{shiomi2014towards}. Ignoring the spacial structured mobile robots with geometric envelopes restricts the volume of permitted workspace and limits the close interaction between mobile robots and obstacles in a clustered environment.

 When considering the spatial structure of the mobile robotic arm, the rigid body links are of great interest in literature, where \cite{chang1994collision} approximated the links of the robotic arm by a set of convex polyhedra while \cite{ju2001fast} treated them as convex ellipsoids.
Caused by the high degrees of freedom, calculating the motion control law for the entire mobile robotic arm can be computationally expensive. In \cite{lumelsky1993real}, a hybrid teleoperation system was proposed, where the
operator carries out general control for the mobile base while the collision
avoidance task is shifted to sensor-based robotic arm subsystem.
This decouples the motion constraints between the mobile base and the robotic arm. Therefore, we propose to find a way of incorporating the spatial structure of the mobile robotic arm while conducting collision avoidance task, treating the mobile robotic arm as a whole.

\begin{figure}[t]
    \centering
    \includegraphics[width=8cm]{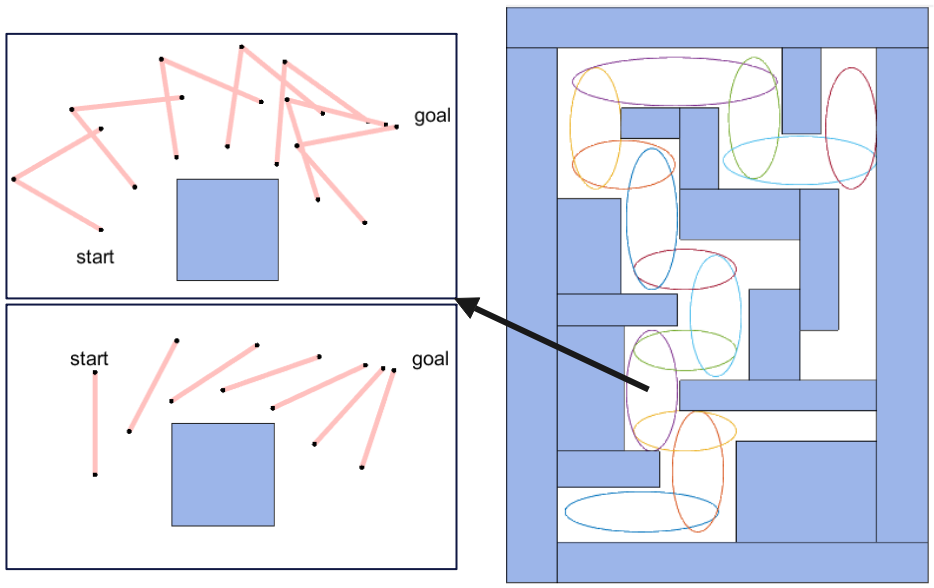}
    \caption{Scenario of task scene and expected realization.
    The maze environment in the right figure serves as the configuration space, in which a series of convex ellipsoids are generated as a connected corridor from the start place to the goal region.
    In the corner, for instance, where the black arrow starts, we expect the rigid rod and the double link bypass the obstacle without collision.
    Note that both the rigid rod and the double link possess rigid body restrictions, which are considered and solved in an elegant way in this paper.
    }
    \label{fig:start}
    \vspace{-5pt}
\end{figure}

Recently, the Control Barrier Functions (CBFs) approach is proposed as a promising and efficient tool for safe control \cite{ames2014control}. By using a quadratic programming based framework the safety requirements can be encoded as affine constraints. The CBFs method is flexible for an arbitrary nominal control law, which determines the desired behavior. Significant success for this approach has been shown in robotics applications, especially in collision avoidance \cite{glotfelter2017nonsmooth}. Extensions such as considering actuation capacity \cite{chen2020guaranteed}, safety-critical Lagrangian systems \cite{barbosa2020provably}, MPC-CBF \cite{2021arXiv210912313T} for collision avoidance have been proposed. Although these methods are shown to be mathematically rigorous, there is a theory-reality gap dual to the spatial structure of both robots and obstacles.

In this paper, we aim at synthesizing a configuration-aware safe control law for a mobile robotic arm
with considered complex spatial structure. 
The proposed safe control law is able to drive the mobile robotic arm to the goal region while avoiding collision in a obstacle-clustered environment. The challenges brought by the spatial structure are solved by ensuring limited edge points of the robot to be within a series of connected convex obstacle-free sets. By solving an online CBFs-based QP with limited number of constraints, the synthesis approach is quite efficient and promising. 
The main contributions are stated as follows:
\begin{itemize}
    \item We synthesize a configuration-aware control law that can successfully drives our designed mobile robotic arm to the desired region while avoiding collision with environmental obstacles.
    \item The proposed approach incorporates the spatial structure of the mobile robotic arm by merging geometric restrictions into CBFs constraints, thus allowing to solve the collision avoidance problem by convex program.
    \item Our approach is proved to scale well with dimension as the degrees of freedom of the mobile robotic arm increases.
\end{itemize}

The remainder of this paper is organized as follows. Section \ref{sec:structure} introduces the physical structure of the mobile robotic arm.
Preliminaries and reach-avoid-grasp task are specified in Section \ref{sec:task}. 
Main results of this paper are shown in Section \ref{sec:cbf}. Extensive numerical simulations are presented in Section \ref{sec:simulation}. We conclude the paper in Section \ref{sec:conclusion}.


\section{Structure of the Mobile Robotic Arm}
\label{sec:structure}

\begin{figure}[htp]
    \centering
    \includegraphics[width=8cm]{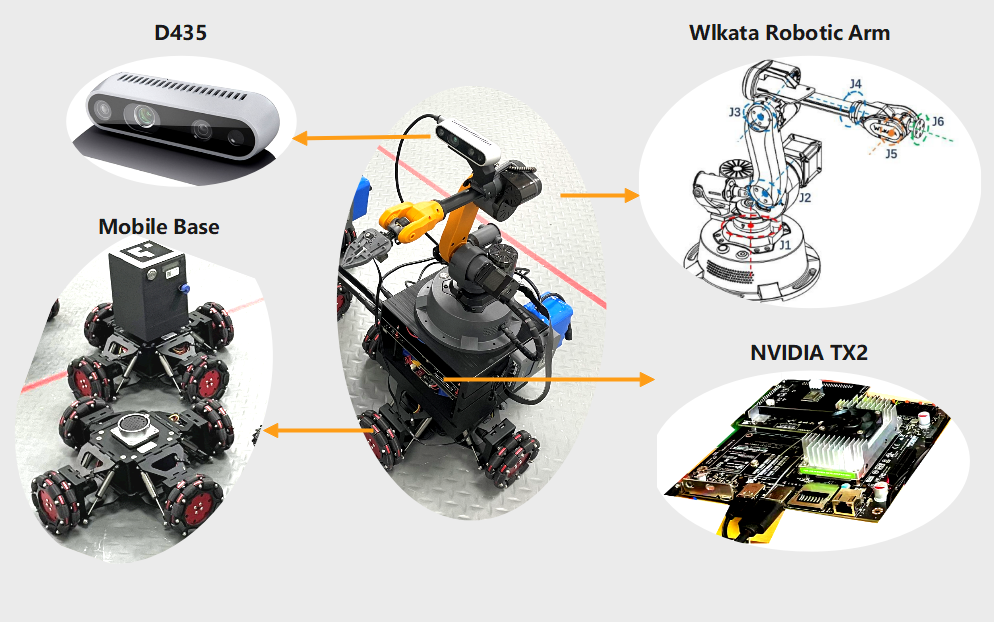}
    \caption{The systematic structure of the mobile robotic arm.}
    \label{fig:robot}
\end{figure}

Our designed mobile robotic arm is shown in Fig. \ref{fig:robot}. It consists of a holonomic four-Mecanum-wheel mobile base and an on-board robotic arm, also equipped with a stereo camera on the top of it. NVIDIA TX2 is embedded as the central computing unit. We tailor a 3D-printed shell as the robot body, on which the robotic arm is set to execute the grasping configuration. The equipped robotic arm is Wlkata Robotic Arm, which possesses the standard structure of industrial manipulators with six (plus one) revolute joints.
The stereo camera (D435) is mounted on the link between joint 3 and joint 4 of the robotic arm, thus constituting the eye-in-hand configuration which provides higher flexibility for robot operations and precise sights of the target \cite{dong2015position}.

\begin{figure}[htp]
    \centering
    \includegraphics[width=8cm]{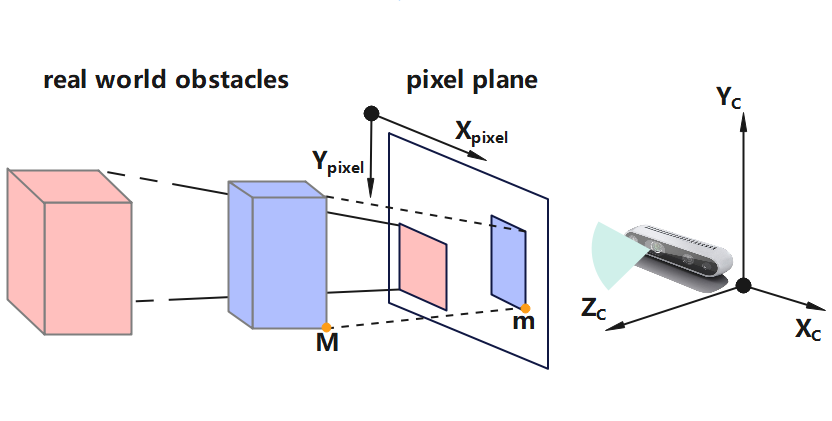}
    \caption{
    Real world reconstruction frame.
    The pink and purple bounding boxes in the pixel plane are the detected results achieved by the HSV-based color segmentation algorithm \cite{shaik2015comparative} and minimum volume bounding box method \cite{huebner2008minimum}.
    The corresponding cubes are the reconstructed real-world obstacles  
    }
    \label{fig:polygon}
\end{figure}

\begin{figure}[htp]
    \centering
    \includegraphics[width=8cm]{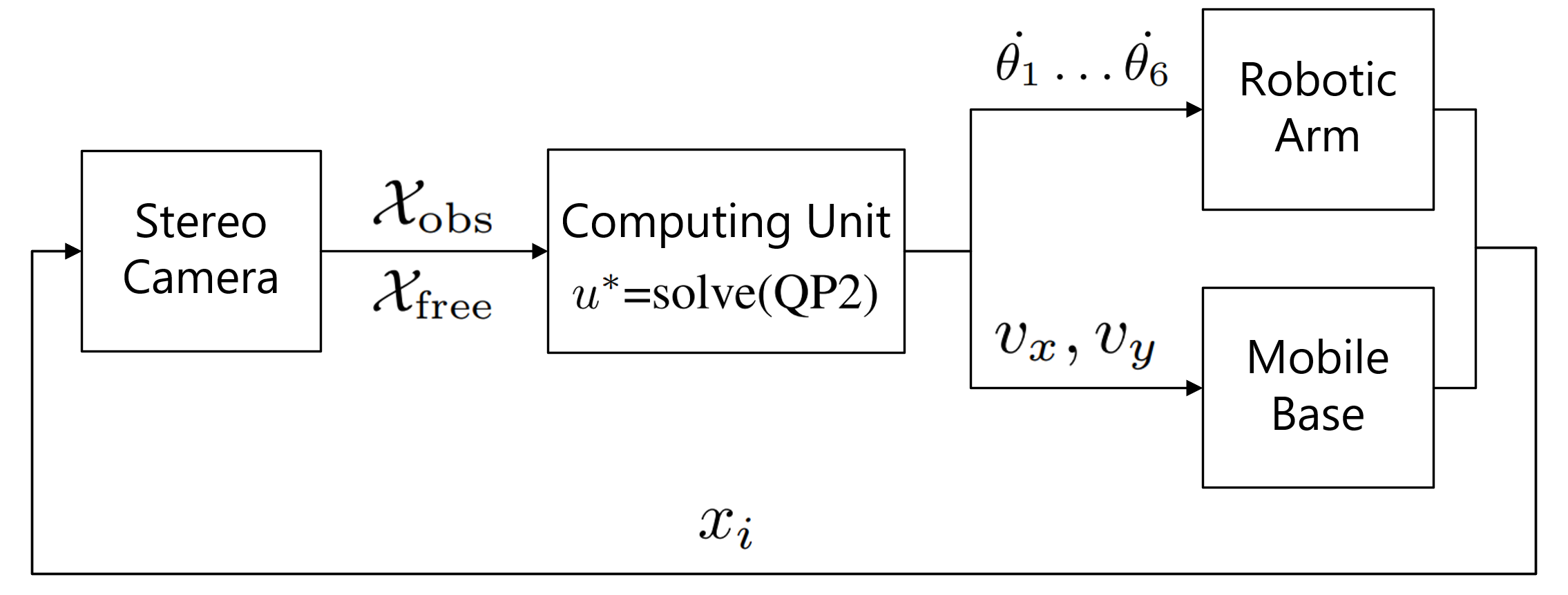}
    \caption{The control framework of our designed mobile robotic arm. 
    The obstacle-free space $\mathcal{X}_\mathrm{free}$ and the obstacle space $\mathcal{X}_\mathrm{obs}$ are approximated with local 3D real world reconstructed by the stereo camera.
    Then the computing unit utilizes the obstacle-free space $\mathcal{X}_\mathrm{free}$ and obstacle space $\mathcal{X}_\mathrm{obs}$ information to solve the control law $u$ from a QP with designed CBF constraints. 
    The control law $u$ comprises of angle velocities $\dot{\theta_1}\ldots\dot{\theta_6}$ of the robotic arm joints and the velocities $v_x,v_y$ of the mobile base.
    The mobile robotic arm state $x_i$ is updated by movement of the mobile base and rotation of the robotic arm and fed back to the stereo camera.
    }
    \label{fig:control frame}
\end{figure}

This stereo camera is an RGB-D camera, from which the RGB information is updated at 30 Hz and the depth information updated at 90 Hz. 
With the Intel RealSense SDK 2.0 (a provided Intel software), the stereo camera can get on-chip self-calibration. 
After that, an HSV (Hue, Saturation and Value)-based color segmentation algorithm \cite{shaik2015comparative} 
and morphology open operation are utilized to detect the colored environmental obstacles.
Then we manage to get reliable depth information $\mathrm{Z_c}$ of the object in the camera coordinate system with a depth information filter \cite{camplani2012adaptive}. 
Let pixel coordinate of point m on the pixel plane be $(\mathrm{X_{pixel}}, \mathrm{Y_{pixel}})$, its real-world coordinate in camera coordinate system $({\mathrm{X_c}}, {\mathrm {Y_c}})$ can be calculated 
according to the following equations
\begin{equation}
\begin{aligned}
\frac{x_\mathrm{pixel}-\mathrm{ppx}}{\mathrm{f_x}}=\frac{\mathrm{X_c}}{\mathrm{Z_c}}   ,\quad
\frac{\mathrm{Y_{pixel}}-\mathrm{ppy}}{\mathrm{f_y}}=\frac{\mathrm{Y_c}}{\mathrm{Z_c}} , \label{XX}
\end{aligned}\nonumber
\end{equation}
where $\mathrm{ppx}, \mathrm{ppy}, \mathrm{f_x},\mathrm{f_y}$ are the camera intrinsic parameters. Local 3D world reconstruction \cite{kagami2005online} frame is performed in Fig. \ref{fig:polygon}.
Thanks to the high-frequency detection rate of the camera, our mobile robotic arm is capable of detecting the surroundings and re-localizing itself in real time. The control flow is illustrated in Fig. \ref{fig:control frame}.

\section{Preliminaries and Problem Formulation}
\label{sec:task}

\subsection{Control Barrier Function}

The set of real numbers is denoted by  $\mathbb{R}$, while  $\mathbb{R}^n$ lifts the dimension to $n$.
Let  $x\in\mathbb{R}^n$ and $u\in\mathbb{R}^m$ be the state vector and control input, respectively. Consider a nonlinear control affine system as
\begin{equation}
   \dot {x} = f(x) + g(x)u,
   \label{eq:model}
\end{equation}
where we suppose that functions $f(\cdot)$ and $g(\cdot)$ are both locally Lipschitz continuous.
Consider a set $\mathcal{C}$ 
defined as the zero-super level set of a
continuously differentiable function $H: \mathbb{R}^n$ $\rightarrow$ $\mathbb{R}$ as

\begin{equation}
   \begin{aligned}
        \mathcal{C}=\{x\in\mathbb{R}^n: H(x)\ge0 \},\\
   \end{aligned}
   \label{eq:C}
\end{equation}
and $\frac{\partial H(x)}{\partial {x}}\neq 0 $ for all $x\in\{x|H(x)=0\}$.

\begin{definition}
A set $\mathcal{C}$ is forward invariant with respect to system (\ref{eq:model}) if for every initial state $x(t_0)\in \mathcal{C}$, there exists $u$, such that the solutions remain within $\mathcal{C}$, i.e., $x(t)\in \mathcal{C} $ for $\forall t\ge t_0$. 
\label{def:forward}
\end{definition}

The system (\ref{eq:model}) is safe with respect to a set $\mathcal{C}$ if $\mathcal{C}$ is forward invariant and $\mathcal{C}\subseteq\mathcal{S}$ where $\mathcal{S}$ is a compact set defined the safe region.
Assuming that $H(x)=0$ at $t=t_0$, a naturally associated condition to keep the set $\mathcal{C}$ forward invariant is $\dot H(x)\ge0$ for any $t\ge t_0$. By applying the chain rule, $\dot H(x)$ is specified as:
\begin{equation}
\begin{aligned}
\dot H(x)=\frac{d H(x)}{d t}
&=\frac{\partial H(x)}{\partial {x}} \frac{\partial {x}}{\partial t}
&=\frac{\partial H(x)}{\partial {x}}\dot {x}.
\end{aligned}\label{eq:dotH}
\end{equation}
Incorporating \eqref{eq:model} into \eqref{eq:dotH}, we have
\begin{equation}
    \frac{\partial H(x)}{\partial {x}}( f(x)+g(x)u )\ge 0,
\end{equation}
which is regarded as the safe constraint. In the real-world applications, this constraint is rather strict that feasible solutions can rarely be obtained and an alternative relaxed formulation was proposed in \cite{ames2014control}.
\begin{definition}
A continuous function $\alpha(\cdot):(-b,a)\rightarrow(-\infty,\infty)$ is said to belong to extended class-$\mathcal{K}$ for some $a,b>0$ if it is monotonicly increasing and $\alpha(0)=0$.
\label{def:K}
\end{definition}

\begin{definition}
\cite{ames2016control} Let $\mathcal{C}\subseteq D \subset \mathbb{R}^n$ be the set defined by $H(\cdot)$. 
Then $H$ is a control barrier function, if there exists an extended class-$\mathcal{K}$ function $\alpha$, such that $\forall x\in D$, $\exists u\in\mathbb{R}^m$, the following inequality holds:
\begin{equation}
    \sup\limits_{u \in \mathbb{R}^m}\Big[\frac{\partial H(x)}{\partial {x}}( f(x)+g(x)u )\Big]\ge -\alpha(H(x)). \nonumber
\end{equation}
\label{def:CBF}
\end{definition}

Definition \ref{def:CBF} renders the set $K_{CBF}$ of control laws $u$ that guarantees the safe set $\mathcal{C}$ forward invariant \cite{ames2019control}. The set of safe control law is defined by:\begin{equation}
    K_\mathrm{CBF}= \Big\{ u\in \mathbb{R}^m: \frac{\partial H(x)}{\partial {x}}( f(x)+g(x)u )\ge  -\alpha(H(x))  \Big\}\label{eq:K}.
\end{equation}

\subsection{Homogeneous Transformation Matrix}
Let $P _M^C\in\mathbb{R}^3$ be the position vector of point $M$ in coordinate $C$ and $P_M^B\in\mathbb{R}^3$ be the position vector of point $M$ in coordinate $B$. If the origins of coordinate $B$ and $C$ coincide, we can find $\textbf{\emph{R}}_B^C\in SO(3)\subset\mathbb{R}^{3\times3}$, such that
\begin{equation}
    P_M^C=\textbf{\emph{R}}_B^C\times P_M^B,\nonumber
\end{equation}
where each column of $R_B^C$ represents the projections of the three axes of coordinate $B$ on each axis of the coordinate $C$. If the origins of coordinate $B$ and $C$ do not coincide, then we can derive a homogeneous transformation matrix $\textbf{\emph{T}}_B^C\in SE(3)\subset\mathbb{R}^{4\times4}$:
\begin{equation}
    \textbf{\emph{T}}_B^C=
    \begin{bmatrix}
    \textbf{\emph{R}}_B^C&p_B^C\\
    0&1
    \end{bmatrix},\label{eq:Tdef}
\end{equation}
where $p_B^C\in\mathbb{R}^3$ denotes the translation offset from the origin of coordinate $B$ to the origin of coordinate $C$, such that
\begin{equation}
\begin{bmatrix}
P_M^C\\
1
\end{bmatrix}=\textbf{\emph{T}}_B^C\times
\begin{bmatrix}
P_M^B\\
1
\end{bmatrix}.
\label{eq:TTT}\nonumber
\end{equation}

\subsection{Reach-Avoid-Grasp Problem Formulation}


The goal of this paper is to find a feasible control law that drives our designed mobile robotic arm to the goal region while avoiding collision with obstacles. 
A region is regarded as the goal region if, in which the mobile robotic arm can execute the grasp configuration, i.e. the mounted robotic arm could grasp the target requiring no more movement of the mobile base.
More specifically, the reach-avoid-grasp problem is posed as follows:

\begin{problem}[Specify Goal Region]
Given a mobile robot workspace $\mathcal{X}\in\mathbb{R}^n$ and a goal position $x_\mathrm{goal}\in\mathcal{X}$ as a priori,
find the goal region $\mathcal{X}_\mathrm{goal}\subset\mathcal{X}$ in which our designed mobile robotic arm can execute the grasp configuration. 
\label{pro:region}
\end{problem}
\begin{problem}[Geometrically Describe the Mobile Robotic Arm]
Find $\mathcal{M}$ proper edge points $a_i$, $i=1\ldots\mathcal{M}$, and assign $\mathcal{M}-1$ line segments $A_i, i=1\ldots\mathcal{M}-1$, where $A_i$ connects 
points $a_i$ and $a_{i+1}$
to represent the mobile robotic arm $\mathcal{A}$:  \[\mathcal{A}=\bigcup_{i=1}^{\mathcal{M}-1}A_i.\]
\label{pro:A}
\end{problem}

\begin{problem}[Find the Safe Control Law]
Given a mobile robot workspace $\mathcal{X}\in\mathbb{R}^n$, and divide it into obstacle-occupied space $\mathcal{X}_\mathrm{obs} \in \mathcal{X}$ and obstacle-free space $\mathcal{X}_\mathrm{free}=\mathcal{X}\setminus\mathcal{X}_\mathrm{obs}$, find a feasible control law that drives the mobile robotic arm $\mathcal{A}$ from initial state $x_0\in\mathcal{X}_\mathrm{free}$ to a state in the goal region $\mathcal{X}_\mathrm{goal}\subset\mathcal{X}_\mathrm{free}$ while avoiding collision with the obstacles, i.e. \[\mathcal{A}(t)\subset\mathcal{X}_\mathrm{free}.\]
\label{pro:u}
\end{problem}

\section{Main Results}
\label{sec:cbf}
\subsection{Goal Region and Grasping}
\begin{figure}[htbp]
\centering
\subfigure[The axis assignment]{\includegraphics[width=4cm]{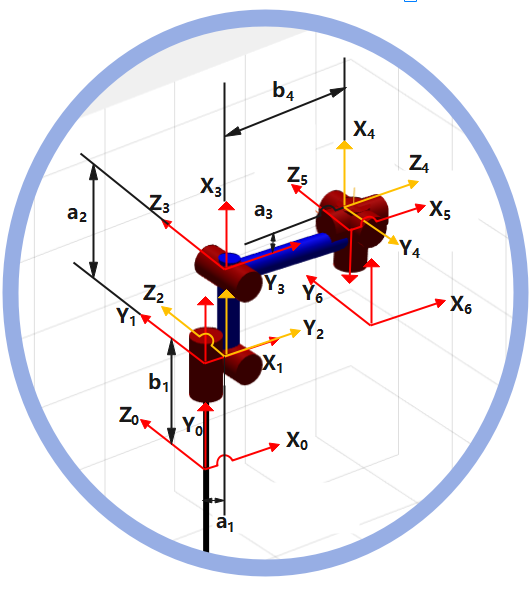}\label{fig:ax}}
\subfigure[The workspace]{\includegraphics[width=4cm]{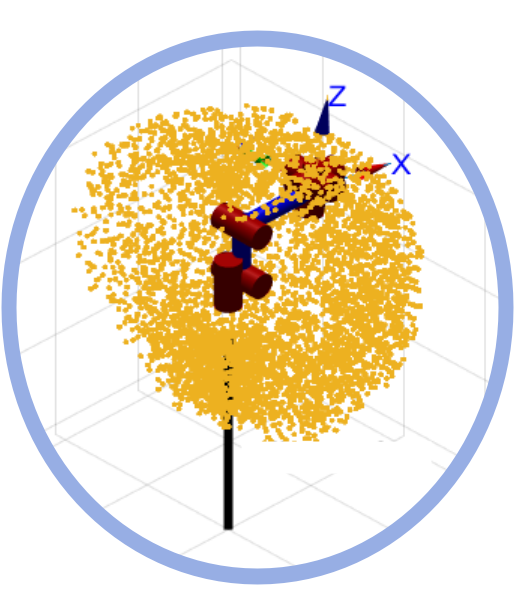}\label{fig:wss}}
\caption{
The axis assignment and workspace of the physical model of our robotic arm. 
In Fig. \ref{fig:ax}, six revolute joints are denoted where blue and red cylinders meet. The blue and red cylinders represent the rigid link between the joints. 
DH parameters $a_1, a_2, a_3$ and $b_1, b_4$ are labeled with double arrows. 
The yellow point cloud in Fig. \ref{fig:wss} numerically describes the workspace of the robotic arm.
Note that the workspace maintains its spatial configuration while translating and rotating with the movement of the robotic arm base.}
\label{fig:axis assignment}
\end{figure}

The key to find the goal region $\mathcal{X}_{goal}$ (i.e. solve Problem \ref{pro:region}) for the mobile robotic arm is to specify the workspace $W$ of the robotic arm. 
The workspace of a robotic arm represents the region where its end-effector can reach without violating the joint limits.
In this paper, we utilize kinematic modeling to numerically describe the workspace of the robotic arm. 
Physical model with  axis assignment is shown in Fig. \ref{fig:ax} and the DH (Denavit-Hartenberg) parameters \cite{farah2013dh} are obtained according to the axis assignment, shown in Table \ref{tab:DH}.
\begin{table}[htbp] 
	\caption{\label{tab:DH}Modified DH model parameters}  
	\centering
		\begin{tabular}{lllllcc} 
 		\toprule 
 	 	coordinate $i$ & $\theta_i$ & $b_i$ & $a_{i-1}$&$\alpha_{i-1}$& Rotation Range \\
  		\midrule 
 		axis1 & 0 & 80&0&0  & -100° to +100° \\ 
 		axis2 & $-\pi$/2 & 0&32&$-\pi$/2& -60° to +90° \\ 
 		axis3 & 0 & 0&108&0 & -180° to +50° \\
 		axis4 & 0 & 176&20&$-\pi$/2& -180° to +180° \\
 		axis5 & $\pi$/2 & 0&0&$\pi$/2& -180° to +40°  \\
 		axis6 & 0 & -20&0&$\pi$/2& -180° to +180°  \\
  \bottomrule 
 \end{tabular} 
\end{table}
The workspace of the robotic arm, which is visualized by yellow point cloud and shown in Fig. \ref{fig:wss}, is obtained by applying the forward kinematics provided by Robotics Toolbox in Matlab. Since the workspace is relevant to the state $x$ of the mobile robotic arm, we rewrite the workspace as $W(x)$. Given the goal position $x_{goal}$, the goal region is defined as
\begin{equation}
    \mathcal{X}_{goal}=\Big\{\boldsymbol x\in\mathcal{X}_{free}:\boldsymbol x_{goal}\in W(\boldsymbol x)  \Big\}.\label{eq:goal}
\end{equation}

\begin{figure}[htbp]
\centering
\subfigure[Relative spatial pose]{\includegraphics[width=4cm]{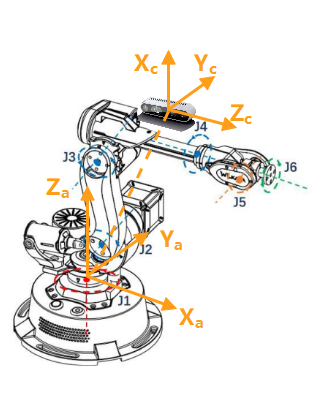}\label{fig:rr1}}
\subfigure[Plane geometric relationship ]{\includegraphics[width=4cm]{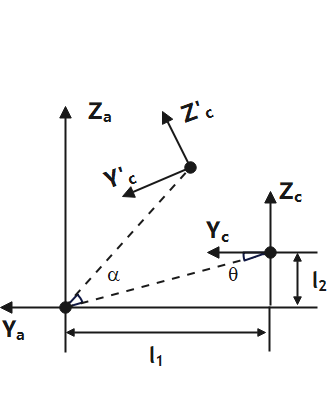}\label{fig:rr2}}
\caption{ 
Geometric relationship between camera coordinate and arm coordinate.
\ref{fig:rr1} shows the relative spatial pose between camera coordinate and arm coordinate on the robotic arm. Joint 1, joint 2, joint 3, joint 4 , joint 5 and joint 6 are abbreviated as J1, J2, J3, J4, J5 and J6.
Geometric relationship between these two coordinates is extracted as \ref{fig:rr2} where axis $\mathrm{Z_c}$ and $\mathrm{Y_c}$ rotate to $\mathrm{Z_c'}$ and $\mathrm{Y_c'}$ by rotation (defined as angle $\alpha$) of J3.
}
\label{fig:r12}
\end{figure}

As for grasping configuration, we consider the homogeneous transformation matrix
\cite{paul1986computationally}, $\textbf{\emph{T}}_c^a$ from the camera coordinate $(X_c Y_c Z_c)$ to
the robotic arm coordinate $(X_a Y_a Z_a)$.
According to \eqref{eq:Tdef}, we have
\begin{equation}
    \begin{bmatrix}X_a& Y_a& Z_a&1\end{bmatrix}^T = \textbf{\emph{T}}_c^a\begin{bmatrix}X_c& Y_c& Z_c&1\end{bmatrix}^T,
    \label{eq:c2a}
\end{equation}
where $\textbf{\emph{T}}_c^a$ is specified as \eqref{eq:T2} calculated with the geometric relationship in Fig. \ref{fig:r12}.
\begin{equation}
\textbf{\emph{T}}_{c}^{a}=
\begin{bmatrix}
1&0&0&0\\
0&\cos\alpha&\sin\alpha&-\sqrt{l_1^2+l_2^2}\cos(\alpha+\theta)\\
0&-\sin\alpha&\cos\alpha&\sqrt{l_1^2+l_2^2}\sin(\alpha+\theta)\\
0&0&0&1
\end{bmatrix}. \label{eq:T2}
\end{equation}
Therefore, when the object is detected by our camera, its coordinate in robotic arm coordinate can be derived from \eqref{eq:c2a} and \eqref{eq:T2}. Then with inverse kinematics tool of Robotics Toolbox in matlab, we can specify the grasp configuration, i.e. angle assignment of each joint.

\subsection{Solution of Reach-Avoid Problem}
The reach-avoid task comprises desired position reaching and obstacle avoiding steps. By deriving the workspace of the robotic arm, the desired position reaching step is relaxed to desired region reaching step, which is less computationally expensive.


We now take advantage of the control framework (shown in Fig. \ref{fig:control frame}) to carry out the reach-avoid task.
When the mobile robotic arm arrives at a position $x_i$, the stereo camera (D435) captures present perspective of surroundings. By applying local 3D world reconstruction method \cite{kagami2005online},
obstacle-free space $\mathcal{X}_\mathrm{free}$ and obstacle-occupied space $\mathcal{X}_\mathrm{obs}$ can be properly approximated. 
Generally speaking, the obstacle-free space is a non-convex region \cite{vasilopoulos2018reactive}, \cite{notomista2021safety}, our approach considers a series of convex sets $\mathcal{C}_0, \mathcal{C}_1,\ldots \mathcal{C}_N$ (i.e. ellipsoids, convex polygons) extracted from the obstacle-free space. 
Based on $\mathcal{X}_\mathrm{free}$ and  $\mathcal{X}_\mathrm{obs}$, proper convex set $\mathcal{C}_i$ enclosing current state $x_i$ can be found by solving convex optimization problems \cite{deits2015computing}.
We assign succeeded state $x_{i+1}$ of $x_{i}$ such that the distance between $x_{i}$ and $x_{i+1}$ is maximized, i.e. solving
\begin{align}\tag{QP1} \label{max:lp2}
\max \quad & \Vert \boldsymbol x_{i}-\boldsymbol x_{i+1}\Vert_2 &{}&\\
\mbox{s.t.}\quad
& \boldsymbol x_{i+1} \in \mathcal{C}_i \subset \mathcal{X}_\mathrm{free},\nonumber
\end{align}




When considering the spatial structure of our designed mobile robotic arm, we mainly focus on the rigid body links (link 1 between J1 and J2, link 2 between J2 and J3, link 3 between J3 and J5, link 4 between J5 and J6 shown in Fig. \ref{fig:rr1}) of the robotic arm. To solve Problem \ref{pro:A},
we choose the central points of J1, J2, J3, J5, J6 as the edge points $a_i$, $i=1\ldots5$.
And abstract the rigid body link of the robotic arm as line segments $A_i, i=1,\ldots,4$, connecting the edge points $a_i$ and $a_{i+1}$,
 to represent the robotic arm $\mathcal{A}$, where
 \begin{equation}
     \mathcal{A}=\bigcup_{i=1}^{4}A_i.\nonumber
 \end{equation}
In a more general sense, we denote $N_l$ as the number of the abstracted line segments and $N_{ep}$ as the number of the edge points. The mobile robotic arm is therefore generalized as:
 \begin{equation}
     \mathcal{A}=\bigcup_{i=1}^{N_{l}}A_i.\nonumber
 \end{equation}
\begin{theorem}
To keep the mobile robotic arm safe in the course of reach-avoid task, we only need to keep the chosen corner points safe, i.e., the safety control is guaranteed by the following control law
\begin{alignat}{2}
u^*(x) = \arg\min_{u \in \cap_{k=1, \cdot, N_{ep}}K^{(k)}_{CBF}} \quad & \Vert u-u^p(x) \Vert_2, &{}& \tag{QP2} \label{eq:finite_}
\end{alignat}
where $u^p(x)=\mathrm{k_p}(x_{i+1}-x_{i})$ and $\mathrm{k_p}$ is the tuned proportional gain.
\label{theo:finite}
\end{theorem}

\begin{proof}
First we prove that the edge points are safe by applying the safe control law. Let $x^{(k)}$ be the state of the edge point k, 
extended from \eqref{eq:K}, we obtain the $K_{CBF}^{(k)}$ with respect to $x^{(k)}$ as:
\begin{equation}
\begin{aligned}
    K^{(k)}_{CBF}= &\Big\{ u \in \mathbb{R}^m:
\Big[\frac{\partial H(x)}{\partial x}( f(x)+g(x) u)\\
&+\alpha(H(x)) \Big]
    \Big|_{x=x^{(k)}} 
    \ge 0 \Big\}.\label{eq:Kk}
\end{aligned}
\end{equation}
Given the states $x^{(k)}$, $x^{(k+1)}$ of edge point $k$, edge point $k+1$, a set $\mathcal{C}$ by \eqref{eq:C} and $u \in K_{CBF}^{(k)}\bigcap K_{CBF}^{(k+1)}$, then $\mathcal{C}$ is ensured forward invariant with \eqref{eq:K} \cite{ames2019control}. 
According to Definition \ref{def:forward}, we can derive that $x^{(k)}$ and $x^{(k+1)}$ are ensured within set $\mathcal{C}$ (i.e. edge point $k$ and edge point $k+1$ are safe).
Then we prove that the mobile robotic arm $\mathcal{A}$ is safe given all the edge points safe.
 For a convex set $\mathcal{C}_i\subset\mathbb{R}^n$, $\forall$ $y^{(1)},y^{(2)}\in\mathcal{C}_i$, equation
\begin{equation}
    \lambda \boldsymbol y^{(1)}+(1-\lambda) \boldsymbol y^{(2)}\in \mathcal{C}_i, \forall \lambda\in[0,1]\nonumber
\end{equation}
holds according to convex set definition. This indicates that if points $y^{(1)}$ and $y^{(2)}$ are within the convex set, then the the generalized line segment connecting them is ensured in this convex set. Earlier in this subsection, we get a series of convex sets $\mathcal{C}_0,\ldots \mathcal{C}_N$ from obstacle free space $\mathcal{X}_\mathrm{free}\subset\mathbb{R}^n$ and now we substitute $\mathcal{C}$ with one of the convex set.
Therefore, given $a_i\in\mathcal{C}, i=1,\ldots,N_{ep}$ (edge points are safe), line segments $A_i\subset\mathcal{C}$(safe), $i=1,\ldots\,N_l$ (line segments are safe), hence $\mathcal{A}=\bigcup_{i=1}^{N_{l}}A_i\subset \mathcal{C}$ (the mobile robotic arm is safe).


\end{proof}




Based on the extracted convex set $\mathcal{C}_0,\ldots \mathcal{C}_N$, the corresponding control barrier function $H_0(x),\ldots,H_N(x)$ can be specified by 
\begin{equation}
    \mathcal{C}_i=\{\boldsymbol x\in\mathbb{R}^n:H_i(\boldsymbol x)\ge0\}, i=0,\ldots, N.
    \label{eq:hi}\nonumber
\end{equation}
The corresponding safe control law $u^*(x)$ is computed by solving \eqref{eq:finite_} with substituting $H(x)$ by the specific $H_i(x)$, where $i$ depends on the current located safe set. For now we have solved Problem \ref{pro:u}.

\section{Numerical Simulation}
\label{sec:simulation}
In this section, simulation of a rigid rod and a mobile robotic arm in a complex maze is performed to verify the proposed algorithm. The QP solver is CVXGEN \cite{mattingley2012cvxgen}.



\begin{figure}[htbp]
\centering
\subfigure[$l=1$]{\includegraphics[width=2.6cm]{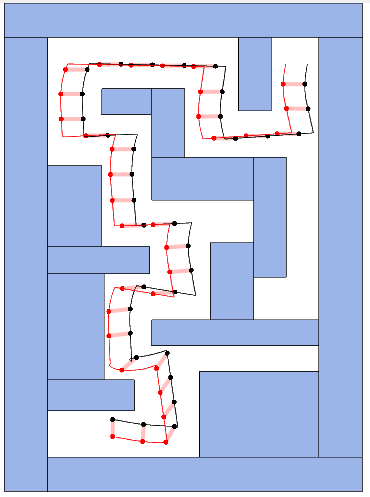}\label{fig:1_}}
\subfigure[$l=1.4$]{\includegraphics[width=2.6cm]{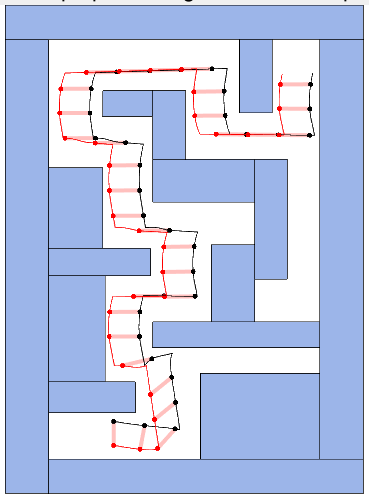}\label{fig:1_4}}
\subfigure[$l=1.8$]{\includegraphics[width=2.6cm]{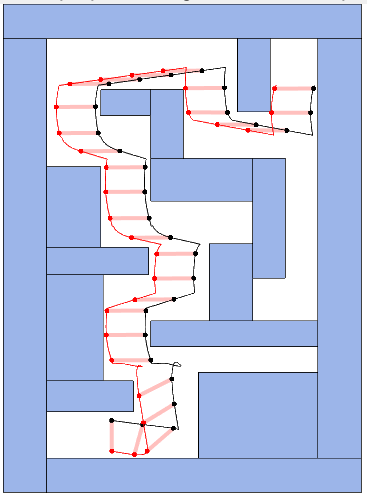}\label{fig:1_8}}

\caption{ Simulation results of a rigid rod in a maze environment with length $l=1$ in Fig. \ref{fig:1_}, $l=1.4$ in Fig. \ref{fig:1_4} and $l=1.8$ in Fig. \ref{fig:1_8}.
The rigid rod (the thick pink segment with a black vertex and a red vertex) aims at navigating from start position to the goal position without collision with obstacles in the maze environment. Slim red curve and black curve are the trajectories of the red vertex and the black vertex. In both cases with different initial pose, the rigid rod successfully and safely reaches the goal region.
}
\label{fig:maze}
\end{figure}

Firstly we apply our method on a rigid rod, aiming at driving the rigid rod from the initial configuration to the goal region without collision with the rectangle obstacles in a maze environment. The maze environment introduces a start position on the bottom left and a goal position on the top right, shown in Fig. \ref{fig:maze}.

\begin{figure}[htbp]
\centering
\subfigure[One joint]{\includegraphics[width=2cm]{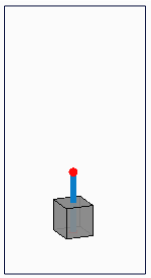}\label{fig:1j}}
\subfigure[Two joints]{\includegraphics[width=2.02cm]{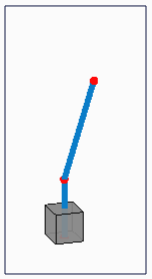}\label{fig:2j}}
\subfigure[Three joints]{\includegraphics[width=1.99cm]{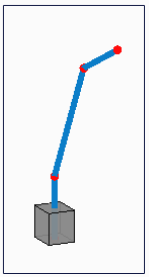}\label{fig:3j}}
\subfigure[Four joints]{\includegraphics[width=2cm]{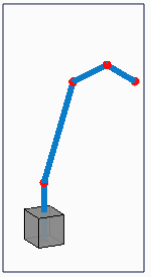}\label{fig:4j}}
\caption{
Models of mobile robotic arm considering different amount of rotating joints. Mobile base is modeled as a grey cube and rigid body links are modeled as blue line segments in Fig. \ref{fig:1j}-\ref{fig:4j}. Red points represent the edge points on the mobile robotic arm, and one edge point is sheltered in the mobile base.
One joint (J1), two joints (J1, J2), three joints (J1, J2, J3) and four joints (J1, J2, J3, J5) are respectively considered in Fig. \ref{fig:1j}, \ref{fig:2j}, \ref{fig:3j} and \ref{fig:4j}.}
\label{fig:jjj}
\end{figure}

Then we conduct simulations on our designed mobile robotic arm.
While describing the workspace of the robotic arm, we notice that J4 and J6 play little role in forming the workspace, both of which are for adjusting the pose of the end-effector to better execute the grasp configuration. 
Therefore, J4 and J6 make little difference in the collision avoidance problem and we focus on J1, J2, J3 and J5 while modeling the mobile robotic arm as shown in Fig. \ref{fig:jjj}.
When four joints are considered, the modeled mobile robotic arm is theoretically close to our designed mobile robotic arm in real world.
According to Theorem \ref{theo:finite}, each joint corresponds to \emph{one} CBF constraint in a convex quadratic programming framework, the algorithm therefore scales well with dimension.
The time consuming of our approach is enumerated as 0.0220$(s)$, 0.0245$(s)$, 0.0279$(s)$ and 0.0292$(s)$ when one, two, three and four joint(s) are respectively considered in our simulation.

Now we present the simulation results of considering three-joint and four-joint mobile robotic arm model.
We refine the state vector $x$ and input vector $u$ in the state space model of the mobile robotic arm as:
\begin{equation}
x=\begin{bmatrix}
     X&
     Y&
     Z
    \end{bmatrix}^T, u=\begin{bmatrix}\dot\theta_1&\dot\theta_2&\dot\theta_3&\dot\theta_5& v_x& v_y\end{bmatrix},^T\label{eq:xxuu}\nonumber
\end{equation}
where $X,Y,Z$  denote the coordinate of the  end-effector , $\dot\theta_1,\dot\theta_2,\dot\theta_3,\dot\theta_5$ are angular velocities of J1, J2, J3, J5 and $v_x,v_y$ are velocities  of the mobile base. 
When specifying the 
state space model of our mobile robotic arm, we only take the most complex situation (four joints) for example, shown as follow 
\begin{equation}
    \begin{split}
    \begin{bmatrix}
    \dot X\\
    \dot Y\\
    \dot Z
    \end{bmatrix}=&
    \begin{bmatrix}
    f_{11}&f_{12}&f_{13}&f_{14}&1&0\\
    f_{21}&f_{22}&f_{23}&f_{24}&0&1\\
    0&f_{32}&f_{33}&f_{34}&0&0
    \end{bmatrix}
    \begin{bmatrix}
    \dot \theta_1\\
    \dot \theta_2\\
    \dot \theta_3\\
    \dot \theta_5\\
    v_x\\
    x_y\\
    \end{bmatrix}\\
    \end{split}
    \label{eq:mmodel},
\end{equation}
where,
\begin{flalign}
&\begin{aligned}
    f_{11}=&-a_2s_1s_2-d_4s_1c_{23}+d_6s_1c_{235},
    \end{aligned}&\nonumber\\
    &\begin{aligned}
    f_{12}=&-a_2c_1c_2-d_4c_1s_{23}+d_6c_1s_{235},
    \end{aligned}&\nonumber\\
    &\begin{aligned}
    f_{13}=&-d_4c_1s_{23}+d_6c_1s_{235},
    \end{aligned}&\nonumber\\
    &\begin{aligned}
    f_{14}= d_6c_1s_{235},
    \end{aligned}&\nonumber\\
    &\begin{aligned}
    f_{21}=&-a_2c_1s_2+d_4c_1s_{23}+d_6s_1s_{235},
    \end{aligned}&\nonumber\\
    &\begin{aligned}
    f_{22}= a_2s_1c_2-d_4s_1s_{23}+d_6s_1s_{235},
    \end{aligned}&\nonumber\\
    &\begin{aligned}
    f_{23}=&-d_4s_1s_{23}+d_6s_1s_{235},
    \end{aligned}&\nonumber\\
    &\begin{aligned}
    f_{24}= d_6s_1s_{235},
    \end{aligned}&\nonumber\\
    &\begin{aligned}
    f_{32}=&-a_2s_2-d_4c_{23}+d_6c_{235},
    \end{aligned}&\nonumber\\
    &\begin{aligned}
    f_{33}=&-d_4c_{23}+d_6c_{235},
    \end{aligned}&\nonumber\\
    &\begin{aligned}
    f_{34}= d_6c_{235}.
    \end{aligned}&\nonumber
\end{flalign}
($s_1$ denotes $\sin\theta_1$, $c_1$ denotes $\cos\theta_1$ and $s_{23}$ denotes $\sin(\theta_2+\theta_3$). $a_2, d_1, d_4, d_6$ are DH parameters in Table \ref{tab:DH}.)

\begin{figure}[htbp]
 \centering
  \subfigure[Simulating process of three-joint mobile robotic arm]
 {
  \begin{minipage}{8cm}
   \centering
   \includegraphics[width=1\textwidth]{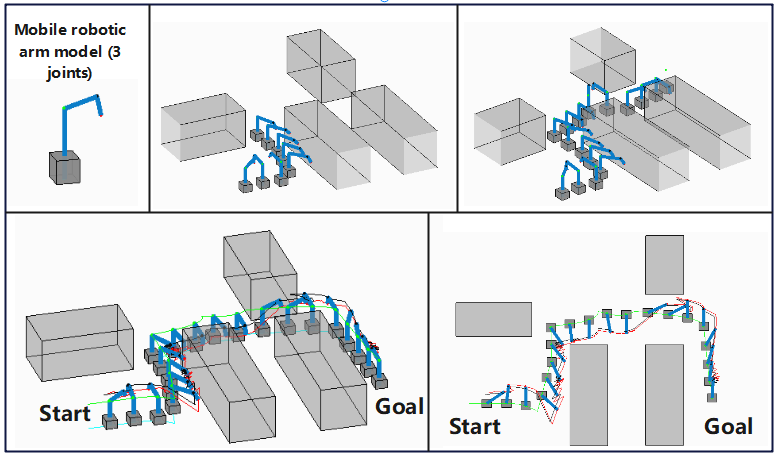}
   \label{j:33}
  \end{minipage}
 }
    \subfigure[Simulating process of four-joint mobile robotic arm]
    {
     \begin{minipage}{8cm}
      \centering
      \includegraphics[width=1\textwidth]{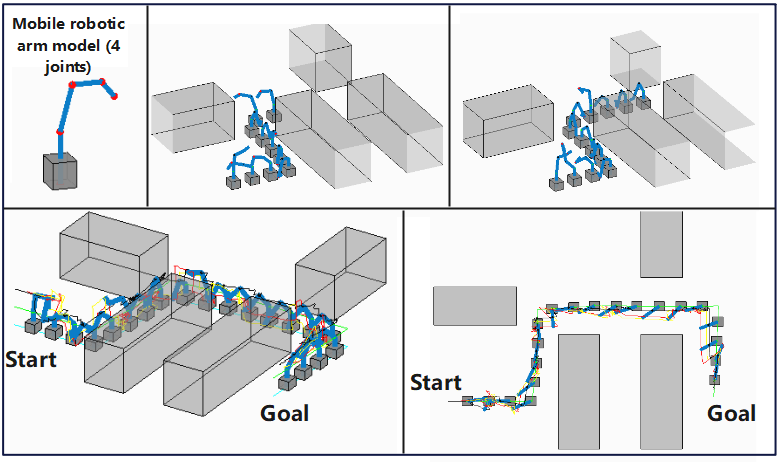}
      \label{j:44}
     \end{minipage}
    }
\caption{Simulating process overview of the modeled mobile robotic arm. By applying the safe control law solved by \eqref{eq:finite_}, the mobile robotic arm manages to reach the goal region while avoiding collision with environmental obstacles both in Fig. \ref{j:33} (three-joint mobile robotic arm) and Fig. \ref{j:44} (four-joint mobile robotic arm).
}
\label{fig:3dd}
\end{figure}

\begin{figure}[htbp]
 \centering
 \subfigure[three-joint mobile robotic arm simulation result]
 {
  \begin{minipage}{7cm}
  \centering
  \includegraphics[width=1\textwidth]{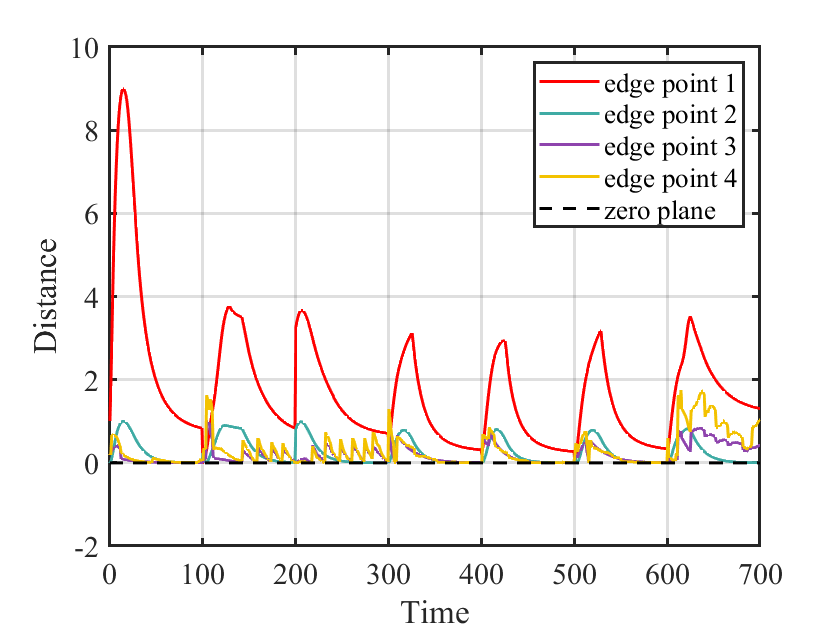}\label{dis:4}
  \end{minipage}
 }
    \subfigure[four-joint mobile robotic arm simulation result]
    {
     \begin{minipage}{7cm}
      \centering
      \includegraphics[width=1\textwidth]{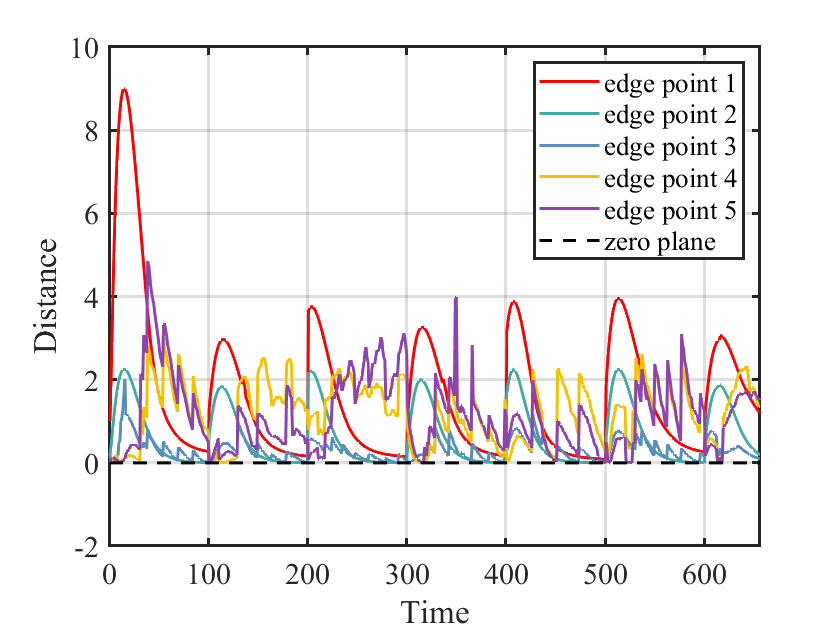}\label{dis:5}
     \end{minipage}
    }
\caption{The minimum distance to the safe set boundary of 4 edge points (see Fig. \ref{dis:4}) and 5 edge points (see Fig. \ref{dis:5}) with respect to time are depicted as curves of different colors. Even though the edge point of the mobile robotic arm may get close to the safe set boundary, the edge point is still ensured safe without collision with the obstacles, also shown in the top view figure in Fig. \ref{fig:3dd}.}
\label{fig:diss}
\end{figure}

The coordinates of the rest edge points can be calculated according to the rigid body restrictions of the mobile robotic arm.  
Driven by the safe control law solved by \eqref{eq:finite_}, our mobile robotic arm can successfully avoid collision with obstacles and navigate to the desired region in the environment where several obstacles are randomly set,
shown in Fig. \ref{fig:3dd}.
Fig. \ref{fig:diss} introduces the minimum distance to the safe set boundary as the indicator of safety and the edge points are ensured in the safe set, hence ensuring the whole mobile robotic arm in the safe set as depicted in Theorem \ref{theo:finite}.


\section{Conclusion and Future Direction}
\label{sec:conclusion}
In this paper, we solved the problem of configuration-aware safe control for a designed mobile robotic arm with Control Barrier Functions.
We firstly propose to geometrically describe our designed mobile robotic arm and mean to fully cooperate the spatial structure of the mobile robotic arm.
Then we merge these geometric restrictions into CBFs constraints and solve the safe control law from a QP with designed CBFs constraints.
This safe control law can successfully navigate the modeled mobile robotic arm to a desired region while avoiding collision with environmental obstacles. 
We show that our method is promising and computationally efficient by conducting numerical simulations.
In the future we will apply our approach to physical test-bed and explore higher dimensional arms.
\bibliographystyle{ieeetr}
\bibliography{ref.bib}
\end{document}